\documentclass[journal]{IEEEtran}

\ifCLASSINFOpdf
\else
	\usepackage[dvips]{graphicx}
\fi
\usepackage{url}
\usepackage{multirow}
\usepackage{graphicx}
\usepackage{amsmath}
\usepackage[colorlinks,bookmarksopen,bookmarksnumbered,citecolor=blue, linkcolor=blue, urlcolor=blue]{hyperref}
\usepackage{bm}
\usepackage{amsfonts}

\begin{document}

\title{Vision Mamba Distillation for Low-resolution Fine-grained Image Classification}

\author{Yao Chen, Jiabao Wang, Peichao Wang, Rui Zhang, and Yang Li
\thanks{Manuscript submitted July 1, 2024. This work has been supported by the Natural Science Foundation of Jiangsu Province under Grant BK20200581.}
\thanks{The authors are with Army Engineering University of PLA, Nanjing 210007, China. (e-mail: chenyao\_19@163.com; jiabao\_1108@163.com; tester\_choser@163.com; 3959966@qq.com; solarleeon@outlook.com;).}
\thanks{Corresponding author: R. Zhang}}

\markboth{IEEE SIGNAL PROCESSING LETTERS, Vol. XX, No. XX, July 2024}
{Chen \MakeLowercase{\textit{et al.}}: Vision Mamba Distillation for Low-resolution Fine-grained Image Classification}
\maketitle
\begin{abstract}
Low-resolution fine-grained image classification has recently made significant progress, largely thanks to the super-resolution techniques and knowledge distillation methods. However, these approaches lead to an exponential increase in the number of parameters and computational complexity of models. In order to solve this problem, in this letter, we propose a Vision Mamba Distillation (ViMD) approach to enhance the effectiveness and efficiency of low-resolution fine-grained image classification. Concretely, a lightweight super-resolution vision Mamba classification network (SRVM-Net) is proposed to improve its capability for extracting visual features by redesigning the classification sub-network with Mamba modeling. Moreover, we design a novel multi-level Mamba knowledge distillation loss boosting the performance, which can transfer prior knowledge obtained from a High-resolution Vision Mamba classification Network (HRVM-Net) as a teacher into the proposed SRVM-Net as a student. Extensive experiments on seven public fine-grained classification datasets related to benchmarks confirm our ViMD achieves a new state-of-the-art performance. While having higher accuracy, ViMD outperforms similar methods with fewer parameters and FLOPs, which is more suitable for embedded device applications. Code is available at \href{https://github.com/boa2004plaust/ViMD}{Github}.
\end{abstract}
\IEEEpeerreviewmaketitle

\section{Introduction}
\label{intro}
\IEEEPARstart{F}{ine-grained} visual classification (FGVC) aims to classify fine-grained sub-categories within a coarse-grained category~\cite{wei}, such as birds~\cite{cub}, cars~\cite{car}, and dogs~\cite{dog}. The existing representative fine-grained classification methods achieve high accuracy by using high-resolution (HR) images containing many informative details as inputs~\cite{CAMF}. However, in real-world applications, images are often captured from large stand-off distances, which makes the region of interest low resolution (LR). These LR object images are usually difficult to classify correctly because they lack the discriminative details of the object. The performance of a well-trained model on the HR fine-grained image dataset will be significantly degraded when applied to the LR FGVC tasks~\cite{LRE}.

\begin{figure}
	\centering{\includegraphics[width=0.5\columnwidth]{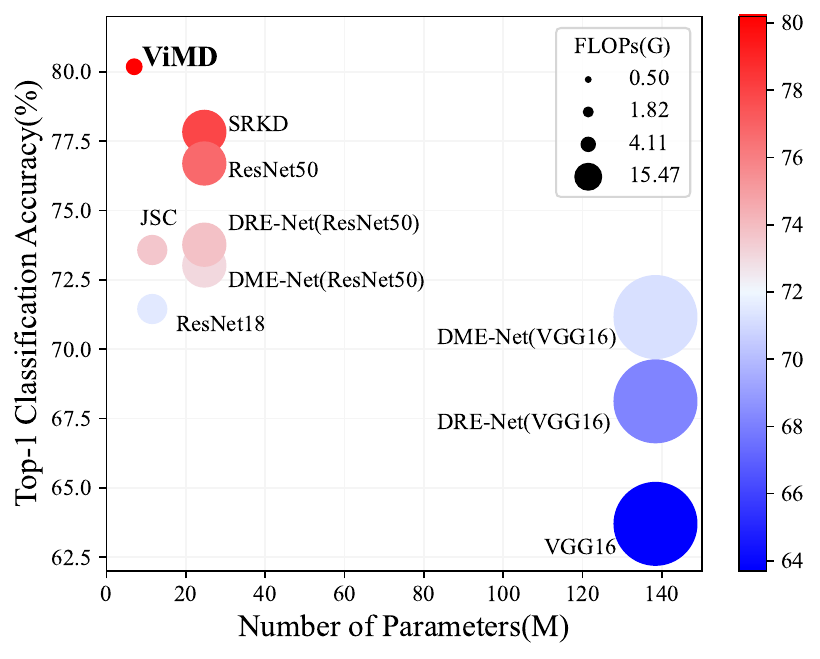}}
	\caption{Effectiveness and efficiency comparison between our ViMD and other methods on Caltech-UCSD Birds 200~\cite{cub} benchmark. The
	color and circle size indicate the model's accuracy and floating point operations,respectively. }
	\label{1}
\end{figure}

To address the challenges associated with LR FGVC, detailed information extracted from HR images is employed to guide the network training on LR images. Depending on the mode used for guidance, current methods can be classified into two categories: super-resolution (SR)-based approaches~\cite{cai,DME,DRE} and knowledge distillation (KD)-based approaches~\cite{SRKD,ASKD,DFD,JSC,FMD,PD}. SR-based approaches can significantly enhance the accuracy of LR FGVC tasks by utilizing SR techniques~\cite{SRGAN,SwinIR} to recover details under the supervision of HR images. Nevertheless, the SR sub-network enlarges the input image size, which increases the parameters and computational cost of the classification sub-network. Consequently, these methods become difficult to be applied to mobile devices with limited resources.

KD-based approaches aim to transfer knowledge from teacher networks pre-trained on the HR images to student networks training on the LR images. To achieve high accuracies, some of these methods~\cite{SRKD,ASKD,DFD} utilize the same structure~\cite{resnet,vgg} for both teacher and student networks, which have ample parameters and computations. Meanwhile, alternative methods~\cite{JSC,FMD,PD} adopt lightweight CNNs~\cite{resnet,shufflenetv2} as students, reducing computational consumption. However, the experimental results of all these methods show that there are still significant gaps in the accuracy of student networks compared to teacher networks.

This letter proposes a \underline{Vi}sion \underline{M}amba \underline{D}istillation (ViMD) method for LR FGVC, which can effectively bridge the gap between the lightweight student networks and the teacher HR networks. Our method can transfer multiple validated Mamba knowledge~\cite{Mamba,vim,vmamba} obtained from teachers into students, which exhibits superior performance~\cite{umamba,detectionmamba,fgvcmamba} compared to traditional CNNs~\cite{resnet} and Transformers~\cite{vit}. In addition, we propose an SRVM-Net to improve its capability for extracting visual features with Mamba modeling, and further design a novel multi-level Mamba knowledge distillation loss to guide the SRVM-Net training with high-quality Mamba knowledge from teacher HRVM-Net.

\begin{figure*}
	\centering{\includegraphics[width=2.0\columnwidth]{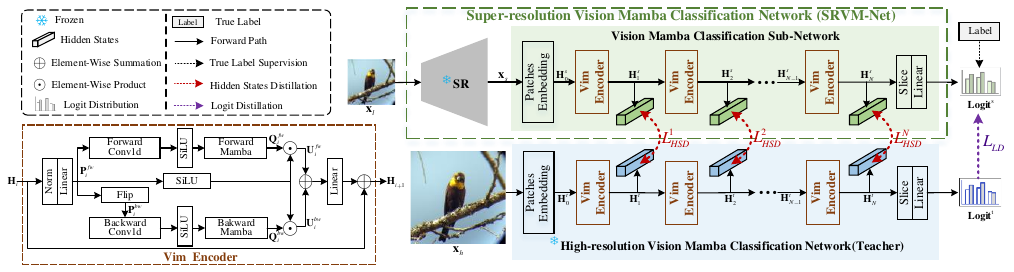}}
	\caption{An overview of ViMD, which is mainly composed of an SRVM-Net (student), an HRVM-Net (teacher), and a multi-level Mamba knowledge distillation loss composed of $L_{HSD}$ and $L_{LD}$. In the training phase, HRVM-Net is firstly trained on HR images, and then SRVM-Net is trained on LR images under the supervision of the multi-level Mamba knowledge distillation loss with the help of HRVM-Net. In the test phase, only SRVM-Net is employed, and it can directly output the prediction results when LR images are given.}
	\label{2}
\end{figure*}

The contributions are summarized as follows:

\begin{itemize}
  \item We propose a novel Vision Mamba Distillation method named ViMD, which is a hybrid lightweight Mamba for low-resolution fine-grained image classification.
  \item We design a novel multi-level Mamba knowledge distillation loss, which transfers logits and hidden states knowledge from HRVM-Net into SRVM-Net.
  \item On seven public fine-grained datasets, the experimental results in Fig. \ref{1} show that our proposed method achieves higher accuracies than other SOTA methods.
\end{itemize}

\section{Method}
The architecture of ViMD is shown in Fig. \ref{2}. It includes an SRVM-Net (student), an HRVM-Net (teacher), and a multi-level Mamba knowledge distillation loss. Given an LR image and its corresponding HR image, they are inputted into SRVM-Net and HRVM-Net, respectively. The SRVM-Net is trained under the supervision of the multi-level Mamba knowledge distillation loss by transferring knowledge from the teacher, HRVM-Net.

\subsection{Super-resolution Vision Mamba Network}
\label{SRVM-Net}
The SRVM-Net is constituted by an SR sub-network and a ViM classification sub-network in sequence. For a LR image ${\mathbf{x}_l} \in {{\mathbb{R}}^{{C_l} \times {H_l} \times {W_l}}}$, where ${C_l}$, ${H_l}$ and ${W_l}$ denote channel, height, and width of $\mathbf{x}_l$ respectively, the SR sub-network reconstructs $\mathbf{x}_l$ into a SR image ${\mathbf{x}_s} \in {{\mathbb{R}}^{{C_s} \times {H_s} \times {W_s}}}$, where $C_s$, ${H_s}$ and ${W_s}$ represent channel, height, and width of $\mathbf{x}_s$ respectively. Its function is to restore the detailed information of $\mathbf{x}_l$. To achieve high-quality image reconstruction, the generator of pre-trained SRGAN~\cite{SRGAN} is directly employed as our SR sub-network, and its details can be found in references~\cite{SRGAN,SRKD,JSC}. The ViM classification sub-network is used to classify the generated SR image $\mathbf{x}_s$, which is trained under the supervision of the multi-level Mamba knowledge distillation loss.

Considering the computational cost and publicly available research on visual space state models, Vision Mamba Tiny (Vim-Tiny)~\cite{vim} is selected as the classification sub-network. Of course, other Vision Mamba models such as Vim-Small~\cite{vim} and VMamba~\cite{vmamba} are also applicable. The ViM classification sub-network mainly consists of a Patches Embedding Module, $N$-layers Vision Mamba Encoder, and a Classification Head.

\subsubsection{Patches Embedding Module} As the original Mamba~\cite{Mamba} was designed for 1D sequence, Patches Embedding Module was used to convert 2D images to 1D sequence. For the SR image $\mathbf{x}_s$ produced by the SR sub-network, Patches Embedding Module firstly uses a convolutional operation to down-sample it into patches ${\mathbf{x}_a} \in {{\mathbb{R}}^{D \times {K_h} \times {K_w}}}$, as follows:
\begin{equation}\label{eq1}
\mathbf{x}_a = \mathbf{x}_s \otimes \mathbf{W}
\end{equation}
where $ \otimes $ denotes the convolution operation, $\mathbf{W} \in {{\mathbb{R}}^{D \times C \times J \times J}}$ denotes the convolution kernel, $K_h={H_s/J}$, and $K_w={W_s/J}$.

Then, the 1D sequence ${\mathbf{x}_b} \in {{\mathbb{R}}^{Z \times D}}$, where $Z={K_h}{K_w}$, can be obtained by applying a flattening operation and a transpose operation to ${\mathbf{x}_a}$, as follows:
\begin{equation}\label{eq2}
\mathbf{x}_b = (Flatten(\mathbf{x}_a))^T
\end{equation}
where $Flatten(\cdot)$ denotes the flattening operation.

According to ViT\cite{vit} and BERT\cite{BERT}, we embed the class label and the position information into the sequence ${\mathbf{x}_b}$. The final 1D sequence output ${\mathbf{H}_0^s} \in {{\mathbb{R}}^{{(}{Z} + 1{)} \times D}}$ is computed by embedding a class token at the middle of ${{\bf{x}}_b}$ and then adding a position embedding as follows:
\begin{equation}\label{eq3}
{\mathbf{H}_0^s} = [\mathbf{x}_b^1,\mathbf{x}_b^2, \cdots,\mathbf{x}_b^{\frac{Z}{2}}, {\mathbf{x}_{cls}},\mathbf{x}_b^{\frac{Z}{2}+1}, \cdots,\mathbf{x}_b^{Z}] + {\mathbf{x}_{pos}}
\end{equation}
where ${\mathbf{x}_b^i} \in {{\mathbb{R}}^{1 \times D}}$ denotes the $i$-th token in the sequence ${\mathbf{x}_b}$, ${\mathbf{x}_{cls}} \in {{\mathbb{R}}^{1 \times D}}$ denotes the class token, and ${\mathbf{x}_{pos}} \in {{\mathbb{R}}^{\mathrm{(}{Z} + 1\mathrm{)} \times D}}$ denotes the position embedding.

\subsubsection{$N$-layers Vision Mamba Encoder} Given the initial hidden state ${\mathbf{H}_0^s}$, $N$-layers ViM encoder can sequentially extract features at different levels of $N$. For the $i$-th ViM encoder layer ${E}_{i},(1 \le i \le N)$, the output $\mathbf{H}_{i+1}^s$ can be obtained by inputting the hidden state $\mathbf{H}_{i}^s $,
\begin{equation}\label{eq4}
\mathbf{H}_{i}^s = {E_{i}(\mathbf{H}_{i-1}^s)}
\end{equation}
where the structure of each ViM encoder ${E}_{i}$ is consistent.

The ViM encoder is based on the residual structure of bidirectional sequence Mamba~\cite{vim}. First, the input sequence $\mathbf{H}_{i}^s$ is normalized and linearly projected to produce the forward sequence $\mathbf{P}_{i}^{fw}$, and then a reverse operation is performed on $\mathbf{P}_{i}^{fw}$ to produce the backward sequence $\mathbf{P}_{i}^{bw}$:
\begin{equation}\label{eq5}
\mathbf{P}_{i}^{fw} = Linear(Norm(\mathbf{H}_{i}^s))
\end{equation}
\begin{equation}\label{eq6}
\mathbf{P}_{i}^{bw} = Reverse(\mathbf{P}_{i}^{fw})
\end{equation}

Then, a convolution before the Mamba modeling~\cite{Mamba} is applied to $\mathbf{P}_{i}^{fw}$ and $\mathbf{P}_{i}^{bw}$ in order to prevent independent token calculations as follows:
\begin{equation}\label{eq7}
\mathbf{Q}_{i}^{fw} = {{M}_{i}^{fw}}(\sigma(\mathbf{P}_{i}^{fw} \otimes \mathbf{W}_i^{fw}))
\end{equation}
\begin{equation}\label{eq8}
\mathbf{Q}_{i}^{bw} = {{M}_{i}^{bw}}(\sigma(\mathbf{P}_{i}^{bw} \otimes \mathbf{W}_i^{bw}))
\end{equation}
where $ \otimes $ denotes the convolution operation, $\mathbf{W}_i^{fw}$ and $\mathbf{W}_i^{bw}$ denote the convolution parameters for the forward sequence $\mathbf{P}_{i}^{fw}$ and the backward sequence $\mathbf{P}_{i}^{bw}$ in $i$-th layer, $\sigma(\cdot)$ denotes SiLU activation, ${M}_{i}^{fw}(\cdot)$ and ${M}_{i}^{bw}(\cdot)$ denote the forward and backward Mamba computation processes.

Finally, the output of $i$-th ViM encoder, ${\mathbf{H}_{i}^s}$, is obtained by
\begin{equation}\label{eq9}
{\mathbf{H}_{i}^s} = {Linear(\mathbf{U}_{i-1}^{fw} + \mathbf{U}_{i-1}^{bw}) + \mathbf{H}_{i-1}^s}
\end{equation}
where $Linear(\cdot)$ represents a linear projection, the forward sequence $\mathbf{U}_{i-1}^{fw}= \sigma(\mathbf{H}_{i-1}^s) \odot {\mathbf{Q}_{i-1}^{fw}}$ and the backward sequence $\mathbf{U}_{i-1}^{bw} = \sigma(\mathbf{H}_{i-1}^s) \odot {\mathbf{Q}_{i-1}^{bw}}$, and $\odot$ represents element-wise multiplication. The whole equation is the classical residual structure, in which the gradients can be efficiently computed and updated during back-propagation.

\subsubsection{Classification Head} To predict the label for a given image, a linear projection is used to map the class token ${\mathbf{h}_{cls}}$ of $\mathbf{H}_N^s=[\mathbf{h}^1, \cdots,\mathbf{h}^{\frac{{Z}}{2}}, {\mathbf{h}_{cls}}, \mathbf{h}^{\frac{{Z}}{2}+1},\cdots,\mathbf{h}^{{Z}}]$ to $\mathbf{Logit}^s$ by
\begin{equation}\label{eq10}
\begin{split}
\mathbf{Logit}^s = Linear(\mathbf{h}_{cls})
\end{split}
\end{equation}
where $\mathbf{Logit}^s=[p^1, p^2, \cdots, p^{C}]$ is the predicted probabilities for all categories. It is used to computed the objective loss in training stage and also used to predict the class label in testing stage. The prediction result $\mathop {\hat{p}^s}$ is computed by \begin{equation}\label{eq11}
\mathop {\hat{p}^s} = \arg \mathop {\max }\limits_i {\mathbf{Logit}^s(i)}
\end{equation}
where $\mathbf{Logit}^s(i)$ denotes the $i$-th value of $\mathbf{Logit}^s$.

\subsection{Multi-level Mamba knowledge distillation loss}
\label{HRVM-DL}
To improve the generalization ability of our SRVM-Net, we design a multi-level Mamba knowledge distillation loss based on logits and hidden states to supervise the training process. For LR FGVC, it is essential to improve its capability in capturing detailed information from SR images and learning prior knowledge from the network pretrained on HR images. As a result, we build a HRVM-Net as the teacher, which can extract fine-grained hidden states and logits from the HR images. Based on the relevant distillation works on Transformer ~\cite{mopeclip,Tinybert,dynabert} and KD~\cite{KD}, a multi-level Mamba knowledge distillation loss ${L_{MKD}}$ is proposed as
\begin{equation}\label{eq12}
{L_{MKD}} = {L_{LD}} + \beta {L_{HSD}}
\end{equation}
where hyper-parameters $\beta $ is used to balance the two losses, ${L_{LD}}$ represents the logits distillation loss function
\begin{equation}\label{eq13}
{L_{LD}} = KL(softmax(\frac{\mathbf{Logit}^s}{{\varDelta}})||softmax(\frac{\mathbf{Logit}^t}{{\varDelta}}))
\end{equation}
where $KL(\cdot)$ denotes the KL divergence function, $\varDelta$ denotes the temperature of the KL divergence function, $\sigma(\cdot)$ denotes the softmax function, and $\mathbf{Logit}^t$ and $\mathbf{Logit}^s$ denote the the logits of teacher and student. And more, ${L_{HSD}}$ represents the hidden states distillation loss
\begin{equation}\label{eq14}
{L_{HSD}} = \sum\limits_{i = 1}^N {L_{HSD}^i}=\sum\limits_{i = 1}^N {|| \mathbf{H}_i^t - \mathbf{H}_i^s} ||_2^2
\end{equation}
where $||\cdot|{|_2}$ denotes the L2 distance, and ${\mathbf{H}}_i^t (i=1,2, \cdots,N)$ denotes the output of the $i$-th teacher's ViM encoder. Both HRVM-Net and ViM classification sub-network has the same network structure.
The teacher can extracted the fine-grained features from the HR images. The student can reach to the teacher by minimizing both ${L_{LD}}$ and ${L_{HSD}}$.

Besides, the cross-entropy loss is also introduced to supervise the training process. The total loss is computed as
\begin{equation}\label{eq15}
L_{total} = {L_{CE}} + \alpha {L_{MKD}}
\end{equation}
where hyper-parameters $\alpha $ is used to balance the losses, ${L_{CE}}$ represents the cross-entropy classification loss function.

\section{Experiments}
\begin{table*}
\caption{Comparison with the State-of-the-Art Methods}
\label{Tab_1}
\centering
\scalebox{0.73}{
\begin{tabular}{cccccccccccc}
\hline
Method                & Teacher  & Student  & CUB             & CAR                & DOG            & PET            & Flower         & MIT67          & Action   &Params$\downarrow$     &FLOPs$\downarrow$    \\ \hline
HR-HR                 & /        & Vim-Tiny & 84.86           & 89.74              & 88.33          & 94.60          & 96.94          & 82.24          & 86.53    &\multirow{2}{*}{6.99}  &\multirow{2}{*}{0.50}    \\
LR-LR                 & /        & Vim-Tiny & 49.14           & 56.95              & 59.36          & 76.45          & 77.98          & 49.55          & 54.37                                                  \\ \cline{1-12}
\multirow{2}{*}{DRE-Net~\cite{DRE}}    & /        & VGG16 & 68.12           & 82.64              & -              & -              & -              & -              & - &138.36 &15.47\\
& /        & ResNet50 & 73.77           & 86.64              & -              & -              & -              & -              & -        &24.66  &4.11\\
    
\multirow{2}{*}{DME-Net~\cite{DME}}    & /        & VGG16 & 71.16           & 87.82& -              & -              & -              & -              & -          &138.36 &15.47                                            \\
& /        & ResNet50 & 73.02           & {\underline{88.38}}& -              & -              & -              & -              & -       &24.66 &4.11                                               \\     \hline

SRKD~\cite{SRKD}      & ResNet50 & ResNet50 & {\underline{77.84}}& -               & -              & -              & -              & -              & -                 &24.66 &4.11                                     \\ \cline{11-12}
JSC(SRGAN)~\cite{JSC} & ResNet50 & ResNet18 & 73.58           & 88.15              & 73.68          & 87.51          & 87.10          & 69.78          & 72.79    &\multirow{2}{*}{11.54} &\multirow{2}{*}{1.82}\\
JSC(SwinIR)~\cite{JSC}& ResNet50 & ResNet18 & 73.70           & 88.24     & {\underline{73.84}}  & {\underline{87.76}}     & {\underline{87.90}}   & {\underline{70.82}}   & {\underline{73.10}}                \\ \hline
\multirow{2}{*}{ViMD (Ours)}    & /        & Vim-Tiny & 77.98           & 86.59              & 83.51          & 92.29          & 92.62          & 75.37          & 80.50    &\multirow{2}{*}{6.99}  &\multirow{2}{*}{0.50}\\
             & Vim-Tiny & Vim-Tiny & \textbf{80.19}  & \textbf{88.93} & \textbf{84.18} & \textbf{92.56} & \textbf{94.03} & \textbf{78.43} & \textbf{83.66}                                             \\
\hline
\end{tabular}
}
\end{table*}

\subsection{Experimental Setup}\label{A}
\subsubsection{Datasets}
\label{dataset}
To evaluate the effectiveness of our ViMD, the LR images are generated by downsampling HR images, because there is no publicly available dataset specifically for LR FGVC. The experiments are conducted on seven public FGVC datasets, including Caltech-UCSD Birds 200 (CUB)~\cite{cub}, Stanford Cars (CAR)~\cite{car}, Stanford Dogs (DOG)~\cite{dog}, Oxford-IIIT Pet (PET)~\cite{pet}, Oxford-102 Flower (Flower)~\cite{flower}, MIT Indoor Scene Recognition (MIT67)~\cite{mit67}, and Stanford 40 Actions (Action)~\cite{action}. Given an original image, a HR image of size 224$\times$224 is got by random cropping, horizontal flipping and image scaling. The corresponding LR image is obtained by down-sampling the HR image to a size of 56$\times$56 using `bicubic' interpolation~\cite{bicubic}.

\subsubsection{Implementation Details}
\label{setting}
In the training phase, we firstly train the teacher (HRVM-Net) initialized with ImageNet1K pre-trained parameters on HR image trainset. The total epochs is 200, and the initial learning rate is 10$^{-6}$ and is tuned by a simulated annealing cosine scheduler. We employ AdamW optimizer with a batch size of 16 and a momentum of 0.9. And then, we train SRVM-Net using the similar configuration as before. For hyper-parameters, $\varDelta$ is set to 4, and the value of $\alpha$ is 1, the details of $\beta$ can be found in Sec. \ref{hyper-param}.

In the testing phase, the teacher (HRVM-Net) can be removed and only the SRVM-Net can be used. Top-1 classification accuracy is employed as evaluation criterion.

\subsection{Comparison with State-of-the-Art Methods}
\label{results}
To evaluate the effectiveness of our proposed ViMD, it is compared with several SOTA methods on seven popular FGVC datasets. SR based approaches, DRE-Net~\cite{DME} and DME-Net~\cite{DRE} are chosen; KD based approaches, SRKD~\cite{SRKD} and JSC~\cite{JSC} are chosen. The results are shown in Table~\ref{Tab_1}, where both SRKD and JSC (SRGAN) are trained with SRGAN as the SR sub-network, while JSC (SwinIR) is trained with SwinIR as the SR sub-network. The Params and FLOPs denote the number of parameters and the floating point operations of classification sub-networks respectively. The best result and the second-best result are highlighted in \textbf{bold} and \underline{underline} respectively. In Table~\ref{Tab_1}, we also present the results of HR-HR (LR-LR), which refers to the ViM classification sub-network trained and tested on HR (LR) images.

Our ViMD achieves the Top-1 classification accuracies of 80.19\%, 88.93\%, 84.14\%, 92.56\%, 94.03\%, 78.43\%, and 83.66\% on seven datasets, respectively, which are all the best results. Compared with SR-based approaches, our method improves the accuracy by 6.42\% on CUB with respect to DRE-Net, and by 0.55\% on CAR with respect to DME-Net. Compared with KD-based approaches, our method improves the accuracy by 2.35\% on CUB with respect to SRKD, by 0.69\%, 10.34\%, 4.80\%, 6.13\%, 7.61\%, and 10.56\% on other six datasets, respectively, with respect to JSC (SwinIR).

The classification sub-network, Vim-Tiny, with only 6.99M Params (approximately 0.05\% of VGG16, 28.3\% of ResNet50 and 60.5\% of ResNet18) and 0.50G FLOPs (approximately 0.03\% of VGG16, 12.1\% of ResNet50 and 27.4\% of ResNet18), are the best. In summary, our ViMD achieves excellent performance with a lightweight architecture, effectively enhancing classification accuracy while reducing computational cost.

\subsection{Ablation Studies}
\label{ablation}

\begin{table}
\centering
\caption{Results of Components Analysis}	
\label{Tab_2}
\centering
\scalebox{0.6}{
\begin{tabular}{ccccccc}
\hline
\multirow{5}{*}{{Components}} & /      &1           &2          &3          &4          &5               \\
                            & ResNet18 &$\surd$     &           &           &           &                \\
                            & SRVM-Net &            &$\surd$    &$\surd$    &$\surd$    &$\surd$         \\
                            & $L_{LD}$ &            &           &$\surd$    &           &$\surd$         \\
                            & $L_{HSD}$&            &           &           &$\surd$    &$\surd$         \\
\hline
\multirow{7}{*}{{Datasets}} & CUB      & 71.45      & 77.98     & 79.96     & 79.10     & \textbf{80.19} \\
                            & CAR      & 85.00      & 86.59     & 88.57     & 87.50     & \textbf{88.93} \\
                            & DOG      & 73.56      & 83.51     & 83.53     & 83.90     & \textbf{84.18} \\
                            & PET      & 87.35      & 92.29     & 92.32     & 92.39     & \textbf{92.56} \\
                            & Flower   & 79.01      & 92.62     & 92.86     & 93.02     & \textbf{94.03} \\
                            & MIT67    & 69.70      & 75.37     & 75.90     & 75.60     & \textbf{78.43} \\
                            & Action   & 72.20      & 80.50     & 82.18     & 80.98     & \textbf{83.66} \\
\hline
\end{tabular}
}
\end{table}

\subsubsection{Analysis of Components}
\label{components}
As illustrated in Table \ref{Tab_2}, ablation experiments are conducted to evaluate the effectiveness of the SRVM-Net and the multi-level Mamba knowledge distillation loss in our proposed ViMD, where all the comparisons employed generator in pre-trained SRGAN as the SR sub-network and trained with $L_{CE}$. Compared with using ResNet18 as the classification sub-network (Column  1), the accuracies obtained with ViM classification sub-network (Column 2) increased by 6.53\%, 1.59\%, 9.95\%, 5.55\%, 13.61\%, 5.67\%, and 8.30\% on the seven datasets, demonstrating the effectiveness of adopting Vim-Tiny as the classification sub-network. Compared with using SRVM-Net without any knowledge distillation method (Column 2), the accuracies obtained by using both $L_{LD}$ and $L_{HSD}$ (Column 5) increased by 2.21\%, 2.34\%, 0.67\%, 0.27\%, 1.41\%, 3.06\%, and 3.16\% on the seven datasets, demonstrating the effectiveness of the designed multi-level Mamba knowledge distillation loss.

\subsubsection{Analysis of Hyper-parameters}
\label{hyper-param}
\begin{figure}
\centering
\includegraphics[width=0.6\columnwidth]{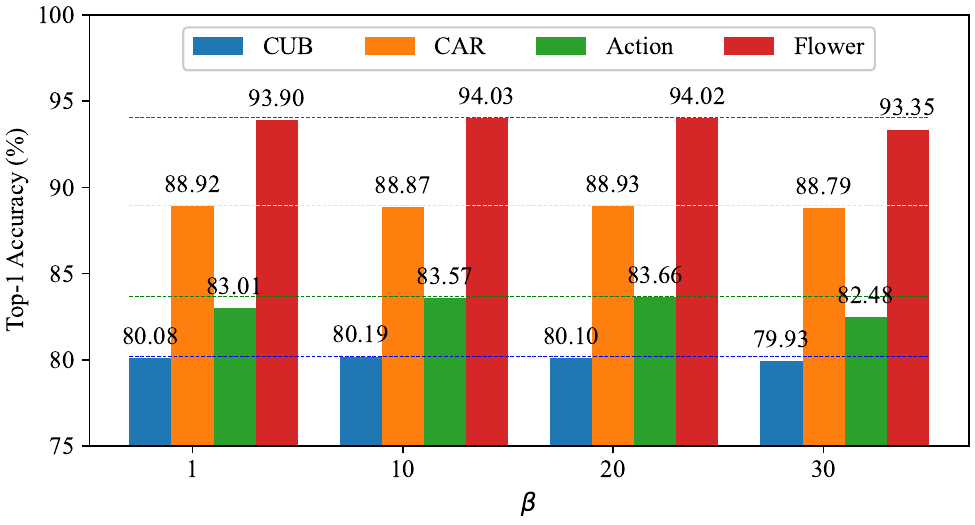}
	\caption{\centering{Results of hyper-parameters analysis on the four datasets.}}
	\label{3}
\end{figure}

Our proposed ViMD uses $L_{CE}$, $L_{LD}$, $L_{HSD}$ together to supervise the training of SRVM-Net. Based on the experience of previous works~\cite{mopeclip,clipkd,catkd}, we set $\alpha$ as 1. To evaluate the affection of different $\beta$, we set $\beta \in \left\{ {1,10,20,30} \right\}$, keeping all other settings as same as Section \ref{setting}. The experimental results are shown in Fig. \ref{3}, where the blue, yellow, green, and red bars respectively indicate the accuracies on CUB, CAR, Action and Flower. It can be found that the performance of our ViMD is robust to hyper-parameter $\beta$, as different $\beta$ perform well for different values. We recommend $\beta$ as 20, because it achieves the best accuracies on CAR and Action. Although it does not achieve the best accuracies on CUB and Flower, the differences between it and the best accuracies are only 0.09\% and 0.01\%, and it still outperforms other SOTA method on all four datasets by 2.26\%, 0.55\%, 7.02\% and 10.56\%.

\section{Conclusion}
In this letter, we propose a Vision Mamba Distillation method for low-resolution fine-grained image classification. The ViMD can effectively balance classification accuracy and computational efficiency with the help of the proposed SRVM-Net and the designed multi-level Mamba knowledge distillation loss. The extensive experiments on seven public fine-grained datasets demonstrate that our ViMD outperforms other SOTA methods, and the effectiveness of its components is verified in the ablation study. We hope this work will inspire further research towards vision mamba and knowledge distillation to boost the performance of LR FGVC tasks.


\newpage

\begin{thebibliography}{39}
\bibliographystyle{unsrt}
\bibitem{wei}X. Wei, {\em et al.}, ``Fine-grained image analysis with deep learning: A survey, '' {\em IEEE Trans. Pattern Anal. Mach. Intell.}, vol. 44, no. 12, pp. 8927-8948, 1 Dec. 2022.
\bibitem{cub}C. Wah, {\em et al.}, ``The caltech-ucsd birds-200-2011 dataset,'' 2011.
\bibitem{car}J. Krause, {\em et al.}, ``3d object representations for fine-grained categorization,'' in {\em Proc. IEEE Int. Conf. Comput. Vis. Workshops}, 2013, pp. 554-561.
\bibitem{dog}A. Khosla, {\em et al.}, ``Novel dataset for fine-grained image categorization,'' in {\em Proc. IEEE Int. Conf. Comput. Vis. Pattern Recognit. Workshop}, 2011.
\bibitem{SRGAN}C. Ledig, {\em et al.},``Photo-Realistic Single Image Super-Resolution Using a Generative Adversarial Network,'' in {\em Proc. IEEE Int. Conf. Comput. Vis. Pattern Recognit.}, 2017, pp. 105-114.
\bibitem{CAMF}Z. Miao, X. Zhao, J. Wang, Y. Li and H. Li, ``Complemental Attention Multi-Feature Fusion Network for Fine-Grained Classification,'' in {\em IEEE Signal Process Lett.}, vol. 28, pp. 1983-1987, 2021
\bibitem{SwinIR}J. Liang, {\em et al.}, ``SwinIR: Image Restoration Using Swin Transformer, '' in {\em Proc. IEEE/CVF Int. Conf. Comput. Vis. Workshops}, 2021, pp. 1833-1844.
\bibitem{cai}D. Cai, {\em et al.}, ``Convolutional low-resolution fine-grained classification, '' in {\em Pattern Recognit. Lett.}, vol. 119, pp. 166-171, Mar. 2019.
\bibitem{DME}T. Yan, {\em et al.}, ``Discriminative Feature Mining and Enhancement Network for Low-Resolution Fine-Grained Image Recognition, '' in {\em IEEE Trans. Circuits Syst. Video Technol.}, vol. 32, no. 8, pp. 5319-5330, Aug. 2022.
\bibitem{DRE}T. Yan, {\em et al.}, ``Discriminative information restoration and extraction for weakly supervised low-resolution fine-grained image recognition, '' in {\em Pattern Recognit.}, vol. 127, pp. 108629, Jul. 2022.
\bibitem{SRKD}H. Chen, {\em et al.}, ``Super-resolution guided knowledge distillation for low-resolution image classification, '' in {\em Pattern Recogn. Lett.}, vol 155, pp. 62-68, Mar. 2022.
\bibitem{ASKD}S. Shin, {\em et al.}, ``Teaching where to look: Attention similarity knowledge distillation for low resolution face recognition, ''in {\em Proc. 17th Eur. Conf. Comput. Vis.}, 2022, pp. 631-647.
\bibitem{JSC}J. Wang, {\em et al.}, ``Low-Resolution Armored Vehicle Identification Approach via Joint Super-resolution and Knowledge Distillation, '' in {\em Journal. Army Engineering University. PLA}, vol 3, pp. 39-47, May. 2024. doi: 10.12018/j.issn.2097-0730.20230505002, [Online].
\bibitem{KD}G. Hinton, O. Vinyals, and J. Dean, ``Distilling the Knowledge in a Neural Network,'' 2015, {\em arXiv: 1503.02531.}
\bibitem{resnet}K. He, {\em et al.}, ``Deep Residual Learning for Image Recognition, '' in {\em  Proc. IEEE/CVF Int. Conf. Comput. Vis.}, 2016, pp. 770-778.
\bibitem{vgg}K. Simonyan, {\em et al.}, ``Very deep convolutional networks for large-scale image recognition, '' in {\em Proc. Int. Conf. Learn. Represent.}, 2015, [Oneline]. Available: http://arxiv.org/abs/1409.1556.
\bibitem{shufflenetv2}N. Ma, {\em et al.}, ``ShuffleNet V2: Practical Guidelines for Effcient CNN Architecture Design, '' in {\em Proc. 15th Eur. Conf. Comput. Vis.}, 2018, pp. 122-138.
\bibitem{Mamba}G. Albert,  {\em et al.}, ``Mamba: Linear-Time Sequence Modeling with Selective State Spaces,'' 2023, {\em arXiv: 2312.00752.}
\bibitem{vim}L. Zhang, {\em et al.}, ``Vision Mamba: Efficient Visual Representation Learning with Bidirectional State Space Model,'' 2024, {\em arXiv: 2401.09417.}
\newpage
\bibitem{vmamba}Y. Liu, {\em et al.}, ``VMamba: Visual State Space Model,'' 2024, {\em arXiv: 2401.10166.}
\bibitem{vit}A. Dosovitskiy, {\em et al.}, ``An Image is Worth 16x16 Words: Transformers for Image Recognition at Scale,'' in {\em Proc. Int. Conf. Learn. Represent.}, 2021, [Online]. Available: https://openreview.net/forum?id=YicbFdNTTy
\bibitem{fgvcmamba}C. Chen, {\em et al.}, ``Res-VMamba: Fine-Grained Food Category Visual Classification Using Selective State Space Models with Deep Residual Learning,'' 2024, {\em arXiv: 2402.15761.}
\bibitem{detectionmamba}T. Chen, {\em et al.}, ``MiM-ISTD: Mamba-in-Mamba for Efficient Infrared Small Target Detection,'' 2024, {\em arXiv: 2403.02148.}
\bibitem{umamba}J. Ma, {\em et al.}, ``U-Mamba: Enhancing Long-range Dependency for Biomedical Image Segmentation,'' 2024, {\em arXiv: 2401.04722.}
\bibitem{pet}O. M. Parkhi, {\em et al.}, ``Cats and dogs, '' in {\em Proc. IEEE Int. Conf. Comput. Vis. Pattern Recognit.}, 2012, pp. 3498-3505.
\bibitem{flower}M. E. Nilsback, {\em et al.}, ``Automated Flower Classification over a Large Number of Classes, '' in {\em Proc. IEEE 6th Indian Conf. Comput Vis., Graph. Image Process.}, 2008, pp. 722-729.
\bibitem{mit67}A. Quattoni and A. Torralba, ``Recognizing indoor scenes, '' in {\em Proc. IEEE Int. Conf. Comput. Vis. Pattern Recognit.}, 2009, pp. 413-420.
\bibitem{action}B. Yao, {\em et al.}, ``Human action recognition by learning bases of action attributes and parts, '' in {\em Proc. IEEE Int. Conf. Comput. Vis.}, 2011, pp. 1331-1338.
\bibitem{BERT}J. Devlin, {\em et al.}, ``BERT: Pre-training of Deep Bidirectional Transformers for Language Understanding, '' 2018, {\em arXiv:1810.04805.}
\bibitem{mopeclip}H. Lin, {\em et al.},``Mope-clip: Structured pruning for efficient vision-language models with module-wise pruning error metric,'' in {\em Proc. IEEE Int. Conf. Comput. Vis. Pattern Recognit.}, 2024, pp. 27370-27380.
\bibitem{Tinybert}X. Jiao, {\em et al.}, ``Tinybert: Distilling bert for natural language understanding, '' 2019, {\em arXiv:1909.10351.}
\bibitem{dynabert}H. Lu, {\em et al.},``DynaBERT: Dynamic BERT with adaptive width and depth,'' in {\em
Adv. Neural Inf. Proces. Syst.}, 2020, pp. 9782-9793.
\bibitem{clipkd}C. Yang, {\em et al.},``CLIP-KD: An Empirical Study of CLIP Model Distillation,'' in {\em Proc. IEEE Int. Conf. Comput. Vis. Pattern Recognit.}, 2024, pp. 15952-15962.
\bibitem{catkd}Z. Guo, {\em et al.},``Class Attention Transfer Based Knowledge Distillation,'' in {\em Proc. IEEE Int. Conf. Comput. Vis. Pattern Recognit.}, 2023, pp. 11868-11877.
\bibitem{bicubic}R. Keys, {\em et al.},``Cubic convolution interpolation for digital image processing,'' in {\em IEEE Trans. Acoust. Speech. Signal Process.}, 1981, pp. 1153-1160.
\bibitem{DFD}M. Zhu, {\em et al.},``Low-resolution Visual Recognition via Deep Feature Distillation,'' in {\em Proc. IEEE Int. Conf. Acoust. Speech. Signal Process.}, 2019, pp. 3762-3766.
\bibitem{FMD}M. Zhu, {\em et al.},``Feature map distillation of thin nets for low-resolution object recognition,'' in {\em IEEE Trans. Image Process.}, 2022, pp. 1364-1379.
\bibitem{PD}M. Zhu, {\em et al.},``Pixel Distillation: Cost-flexible Distillation across Image Sizes and Heterogeneous Networks, '' {\em IEEE Trans. Pattern Anal. Mach. Intell.}, pp. 1-15, 1 July. doi: 2024,10.1109/TPAMI.2024.3421277, [Online].
\bibitem{LRE}Y. Pei, {\em et al.}, ``Effects of Image Degradation and Degradation Removal to CNN-Based Image Classification, '' {\em IEEE Trans. Pattern Anal. Mach. Intell.}, vol. 43, no. 4, pp. 1239-1253, 1 April. 2021.

\end{thebibliography}
\end{document}